\documentclass{article}
\usepackage{spconf,amsmath,graphicx, booktabs, caption}
\usepackage{stfloats}
\usepackage{graphicx}
\usepackage{xcolor}

\makeatletter 
  \newcommand\figcaption{\def\@captype{figure}\caption} 
  \newcommand\tabcaption{\def\@captype{table}\caption} 
\makeatother

\title{A Green Learning Approach to LDCT Image Restoration}

\name{
Wei Wang,
Yixing Wu, 
C.-C. Jay Kuo}
\address{
University of Southern California, Los Angeles, California, USA}

\begin{document}

\maketitle

\begin{abstract}
This work proposes a green learning (GL) approach to restore medical images. Without loss of generality, we use low-dose computed tomography (LDCT) images as examples. LDCT images are susceptible to noise and artifacts, where the imaging process introduces distortion. LDCT image restoration is an important preprocessing step for further medical analysis. Deep learning (DL) methods have been developed to solve this problem. We examine an alternative solution using the Green Learning (GL) methodology. The new restoration method is characterized by mathematical transparency, computational and memory efficiency, and high performance. Experiments show that our GL method offers state-of-the-art restoration performance at a smaller model size and with lower inference complexity.
\end{abstract}


\section{Introduction} \label{sec:intro}

Computed tomography (CT) is a popular imaging tool for medical diagnosis. It uses the multiangular X-ray scanner to collect X-ray beams and reconstructs 3D human internal structures with a series of 2D grayscale images in a non-invasive manner \cite{pan2009computed}. Since the atomic structure of human organs affects the X-ray absorption and attenuation, CT images exhibit tissue characteristics such as shape, dimension, density, etc. \cite{diwakar2018review}. Low-dose CT (LDCT) with a lower radiation dose than normal-dose CT (NDCT) scans is widely adopted in clinics to reduce the impact of radiation. Since a CT image's signal-to-noise ratio (SNR) is proportional to the square root of the amount of X-ray dose \cite{diwakar2018review}, LDCT suffers from a low SNR. The origin of CT image noise can be categorized into two sources \cite{diwakar2018review}. One comes from the X-ray scanning process. There can be statistical noise due to a finite number of X-ray quanta in scanning and random noise in receiving. The other is attributed to electronic devices, which may generate electronic noise and round-off errors in the reception of analog signals. The presence of noise will blur the details and reduce the contrast of CT images and, in turn, affect the precision of the medical diagnosis. LDCT image restoration is an essential topic in medical image processing. 

Numerous traditional methods \cite{wang2006penalized, manduca2009projection, li2014adaptive, feruglio2010block} have been developed for LDCT restoration. Recently, Deep Learning (DL) methods have been proposed because of their excellent capability in image restoration and reconstruction. DL methods \cite{chen2017lowcnn, chen2017low, wolterink2017generative, wang2023ctformer} offer excellent restoration performance but larger model sizes and higher computational costs. 

This work proposes a Green U-shaped Learning (GUSL) method to restore LDCT based on the Green Learning methodology \cite{kuo2023green}. GUSL is a multi-resolution approach that restores LDCT images to NDCT images from coarse to fine resolutions. 
It is characterized by mathematical transparency, computational and memory efficiency, and high performance. We choose five representative DL methods for performance benchmarking. GUSL exhibits competitive restoration performance (in PSNR and SSIM) at a smaller model size and with lower inference complexity.

The rest of the paper is organized as follows. Related work is reviewed in Sec. \ref{sec:review}. The proposed method is detailed in Sec. \ref{sec:method}. Experimental results are presented in Sec. \ref{sec:experiment}. Finally, concluding remarks and future research directions are discussed in Sec. \ref{sec:conclusion}. 

\begin{figure*}[t]
\centering
\includegraphics[trim=10mm 90mm 10mm 90mm, clip=true, width=1.0\linewidth]{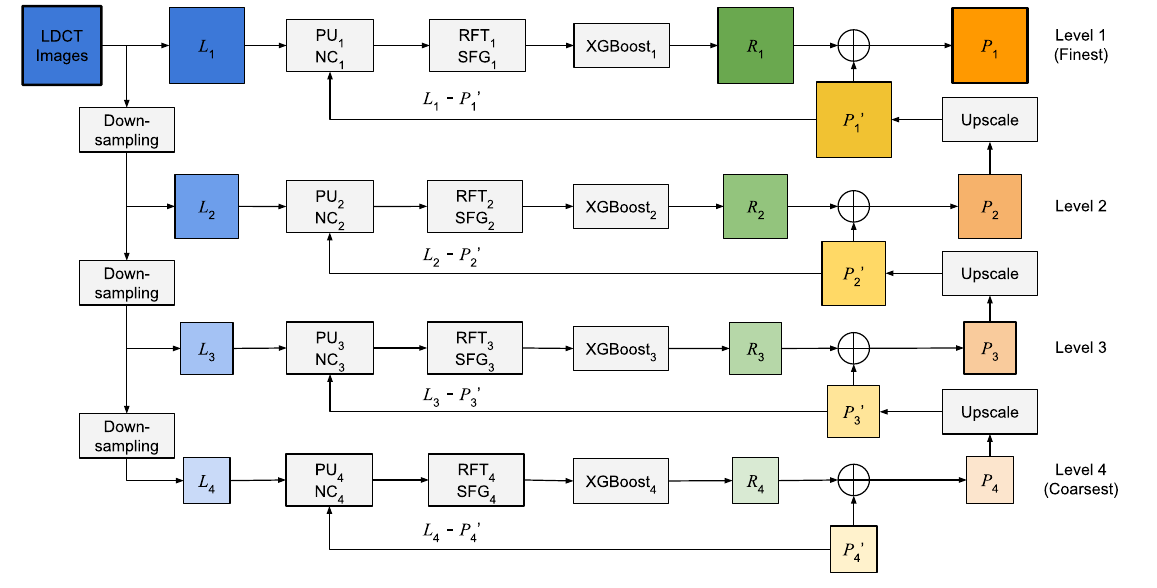}
\caption{The system diagram of the proposed GUSL system. The gray boxes represent the process steps. The colored boxes stand for different types of images, with series of variant spatial sizes and shifting brightness indicating the successively changing resolutions and progressively restored coarse-to-fine information.}\label{fig:pipeline}
\end{figure*}

\section{Review of Related Work} \label{sec:review}

{\bf Traditional Methods.} Three types of conventional methods have been investigated to restore LDCT. The first one performs filtering on raw CT numbers (i.e. the sinogram), such as Penalized Weighted Least Squares \cite{wang2006penalized}, Bilateral filtering \cite{manduca2009projection}, etc. They utilize raw CT data statistics, but the limited source of raw CT data hinders performance. The second is known as iterative reconstruction (IR). They iteratively apply forward and backward projections between the sinogram and image domains to suppress visual noise. However, they are not efficient because of iteration. The third is advanced image restoration techniques such as adaptive non-local means filtering \cite{li2014adaptive}, block matching 3D (BM3D) \cite{feruglio2010block}, etc. Their restored LDCT quality is inferior to that achieved by modern DL methods.

{\bf DL Methods.} DL-based LDCT image restoration has received attention since 2017. Chen {\em et al.} \cite{chen2017lowcnn} proposed a three-layer convolutional neural network, interpreted as sparse-coding, non-linear filtering, and reconstruction. The encoder-decoder network \cite{chen2017low} and GANs \cite{wolterink2017generative} have also been proposed. The transformer \cite{wang2023ctformer} has also been applied. DL methods have state-of-the-art restoration performance. However, the lack of mathematical transparency and high computational complexity are their main weaknesses. 

{\bf Green Learning.} Green Learning \cite{kuo2023green} is an emerging learning paradigm. It has a modular design consisting of three cascaded modules: 1) unsupervised representation learning, 2) supervised feature learning, and 3) supervised decision learning. Green learning has been successfully applied to many applications. The proposed GUSL method follows the general Green Learning pipeline as detailed in Sec. \ref{sec:method}.

\vspace{5pt}
\section{Proposed GUSL Method} \label{sec:method}

\vspace{5pt}
The system diagram of the proposed GUSL system is shown in Fig. \ref{fig:pipeline}. In the training, GUSL uses LDCT images as the input and the corresponding NDCT images as the desired output. GUSL employs a multi-level hierarchical structure to achieve coarse-to-fine sequential restoration. A typical resolution setting at resolution Level $i$ is $h/2^{(i-1)} \times w/2^{(i-1)}$, where $h$ and $w$ are the height and width of the original spatial size. GUSL estimates the residual at each level based on the restoration at the previous level, which is upscaled from the coarser level. The final prediction is obtained by successively upscaling predictions from smaller resolutions plus cumulative residual estimations. This progressive residual correction mechanism efficiently reconstructs fine details from coarser estimates without iteration. Although GUSL shares a similar multi-resolution architecture as U-Net \cite{ronneberger2015u}, it does not have backpropagation required in neuronal network training. All GUSL parameters are determined using a feedforward process under the Green Learning paradigm. At each level, GUSL comprises three steps as described below.

\begin{itemize}
\item Step 1: Gathering Representations as Feature Candidates \\
To gather substantial representations at resolution level $i$, we consider two input sources. One is the downsampled LDCT images $L_i$ obtained by mean-pooling, providing raw LDCT distribution with noise. The other is the difference map between $L_i$ and the upscaled images $P_i'$ interpolated from the restored images $P_{i+1}$ at Level $i+1$, which offers previous correction as long-range spatial guidance. This part facilitates the correction of residuals from coarse to fine levels, thereby enhancing the overall representation. Given a pixel at resolution Level $i$, we process both sources using the PixelHop Unit (PU) to extract joint spatial-spectral representations with $5 \times 5$ and $7 \times 7$ windows centered at that pixel, respectively. Moreover, to capture the correlation between the central pixel and its neighborhood, we conduct neighborhood construction (NC) by including the surrounding $5 \times 5$ pixels and their transform coefficients. Such unsupervised representation learning process ensures a set of comprehensive and robust feature candidates.

\item Step 2: Feature Selection and Generation \\
The raw regression target at resolution Level $i$ is the residual between the downsampled NDCT images at this level and the previous upscaled prediction $P_i'$. We use the residual as supervision and select effective representations using the Relevant Feature Test (RFT) \cite{yang2022supervised}. RFT examines the relevance between a representation dimension and the supervision target through the loss value of RFT. For a certain representation dimension, its range is uniformly segmented into non-overlapping bins, and the central value of each bin is chosen as a candidate threshold. RFT uses one candidate threshold to divide the dataset into two subsets and calculate each partition's weighted mean-squared error (MSE) of the supervised target. The smallest MSE from settings by all thresholds will be assigned as the RFT loss value of the representation dimension, as it captures the potential regression error by treating the mean of the supervision target as the preliminary regression estimation. A lower RFT loss value indicates better discriminative power of the representation. The dimensions before the elbow point in the ascendingly sorted RFT loss curve are then selected as the effective representations. To further increase the robustness of the RFT process, we conduct joint RFT selection. The training data is partitioned into two sets: 80\% for training and 20\% for validation. We perform two independent RFTs on them. A scatter plot is generated where each point represents the ranking in the two datasets. We select the representations near the origin, as these features demonstrate consistent discriminative power in both datasets.

Furthermore, we employ a recently developed Statistics-based Feature Generation (SFG) method \cite{wang2024statistics} to generate more discriminant features. With the supervision of residuals, the SFG method identifies subsets of representations with strong discriminative ability and combines them to generate new features through linear projection. This module comprises two components: (1) representation subset selection using an auxiliary XGBoost model, and (2) new feature generation determined via the Least-Square Normal Transform (LNT). We utilize a relatively shallow XGBoost regressor  \cite{chen2016xgboost} for representation subset selection, allowing it to progressively optimize the partitioning of the target space. By aggregating representations along each tree path, the collected representations form subsets tailored to the corresponding target subspaces. Subsequently, GUSL generates new features by performing a linear combination of representation subsets. The combination weights are determined by a linear projection from each representation subset to the target subspace using LNT. The new features, called LNT features, will be more relevant to the regression target due to the strong supervision of the target.

\item Step 3: XGBoost Regression \\
In the third step, we utilize a XGBoost regressor \cite{chen2016xgboost} to perform residual estimation at each level. At resolution Level $i$, an XGBoost regressor is trained and supervised by the residuals between the ground-truth NDCT images at the current resolution Level $i$ and the interpolated predictions $P_i'$ from the previous resolution Level $i+1$ (Eq. \ref{eq:pred_cross_level}). Using these residuals, XGBoost can refine the restored output by learning and correcting the remaining local fine-grained discrepancies by estimating the conditional residual distribution based on previous predictions. Conditional residuals usually have a relatively smaller range and simpler distribution, thus benefiting the decision process. The restored image prediction $P_i$ at resolution Level $i$ is computed by 
\begin{equation}
P_i = P_i' + R_i,
\label{eq:pred_update}
\end{equation}
where $R_i$ is the estimated residuals and
\begin{equation}
P_i' = \text{Upscale (} P_{i+1} \text{)}
\label{eq:pred_cross_level}
\end{equation}
Prediction at resolution Level $1$ constitutes the final output of the GUSL pipeline, maintaining the same resolution as the original LDCT image.
\end{itemize}

For LDCT images at the coarsest resolution level (i.e., Level 4 in our current case), they are partitioned into multiple $4 \times 4$ patches. We run the K-means algorithm on them and group patches into 1,024 clusters. 
The closest cluster centroid, which is a 16-D vector, serves as the prediction for a certain $4 \times 4$ patch, and the image predictions consisting of $4 \times 4$ patch predictions are the seed prediction $P_4'$.

\begin{figure*}[t]
\centering
\begin{minipage}{1\linewidth} 
\centerline{\includegraphics[width=1.0\linewidth]{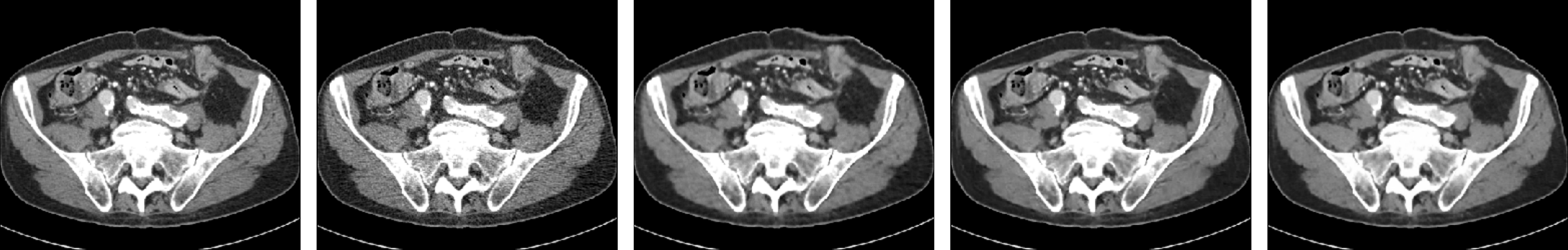}}
\end{minipage}
\begin{minipage}{1\linewidth} 
\centerline{\includegraphics[width=1\linewidth]{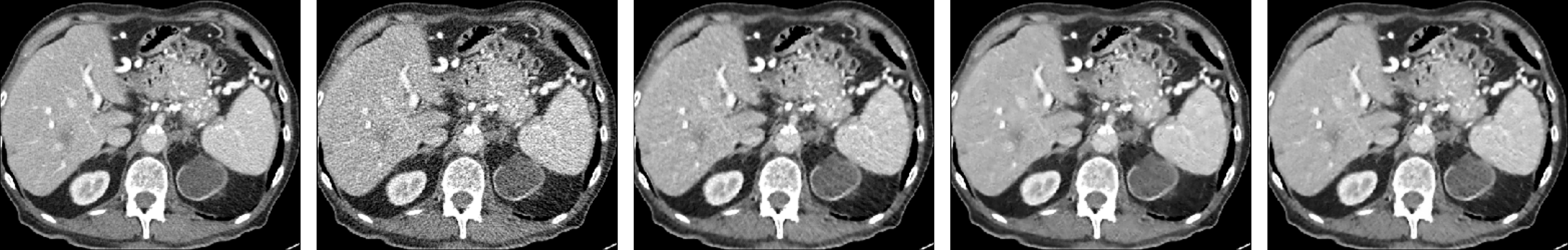}}
\end{minipage}
\begin{minipage}{1\linewidth} 
\centerline{\includegraphics[width=1\linewidth]{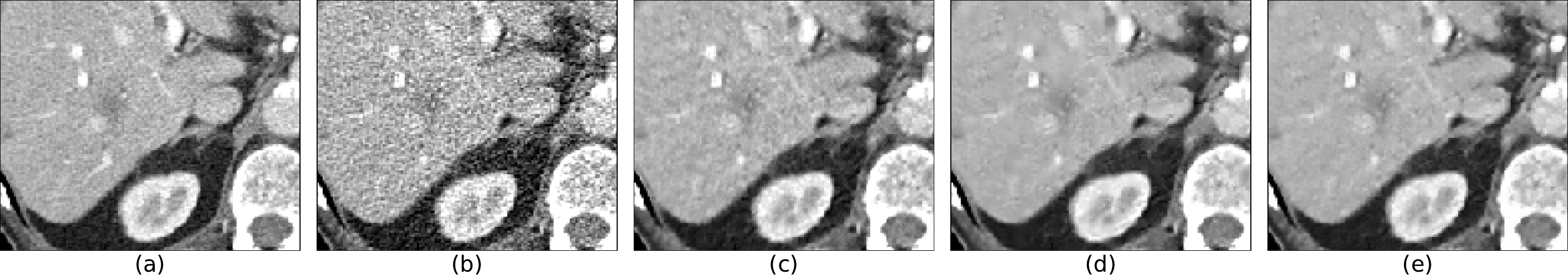}}
\vspace{-5pt}
\end{minipage}
\caption{Visualization comparison of (a) LDCT source images, (b) NDCT source images, (c) RED-CNN restored images, (d) CTformer restored images, and (e) GUSL (ours) restored images.}
\label{fig:visualization}
\vspace{-5pt}
\end{figure*}

\begin{figure*}[t]
\begin{minipage}{1\linewidth} 
\tabcaption{A comparison of GUSL and five benchmarking methods on L506 in terms of PSNR, SSIM, the model parameter count, and the multiplier and addition counts (MACs) per pixel, where red and blue colors indicate the best and the second-best results for each metric.}
\label{tab:comparison}
\vspace{-3pt}
\centering
\footnotesize
\setlength{\tabcolsep}{15pt}{
\begin{tabular}{@{}lcccc@{}}
\hline
\toprule
\textbf{Method} 
    & \textbf{PSNR} & \textbf{SSIM} 
    & \textbf{\#param.} & \textbf{MACs / pixel} \\ 
\midrule
LDCT (Source)    
    & 29.24     & 0.8759  
    & --        & --             \\    
RED-CNN          
    & 32.32     & 0.9053  
    & 1.85M (3.27X)     & 1.23M (38.24X) \\
WGAN-VGG         
    & --        & 0.9008	
    & 34.07M (60.22X)   & 0.88M (27.34X) \\
MAP-NN	         
    & --        & 0.9084	
    & 3.49M (6.17X)	    & 3.37M (104.43X)\\
AD-NET	         
    & --        & 0.9105	
    & 2.07M (3.66X)	    & 2.32M (71.86X) \\
CTformer         
    & \textcolor{red}{\textbf{33.08}}  
    & \textcolor{red}{\textbf{0.9119}}  
    & \textcolor{blue}{\textbf{1.45M (2.56X)}}  
    & \textcolor{blue}{\textbf{0.21M (6.51X)}} \\ 
GUSL (Ours)      
    & \textcolor{blue}{\textbf{33.00}}  
    & \textcolor{blue}{\textbf{0.9111}}  
    & \textcolor{red}{\textbf{0.57M (1X)}}  
    & \textcolor{red}{\textbf{0.03M (1X)}}    \\ 
\bottomrule
\hline
\end{tabular}}
\end{minipage}
\end{figure*}

\begin{figure}[t]
\centering
{\includegraphics[width=0.85\linewidth]{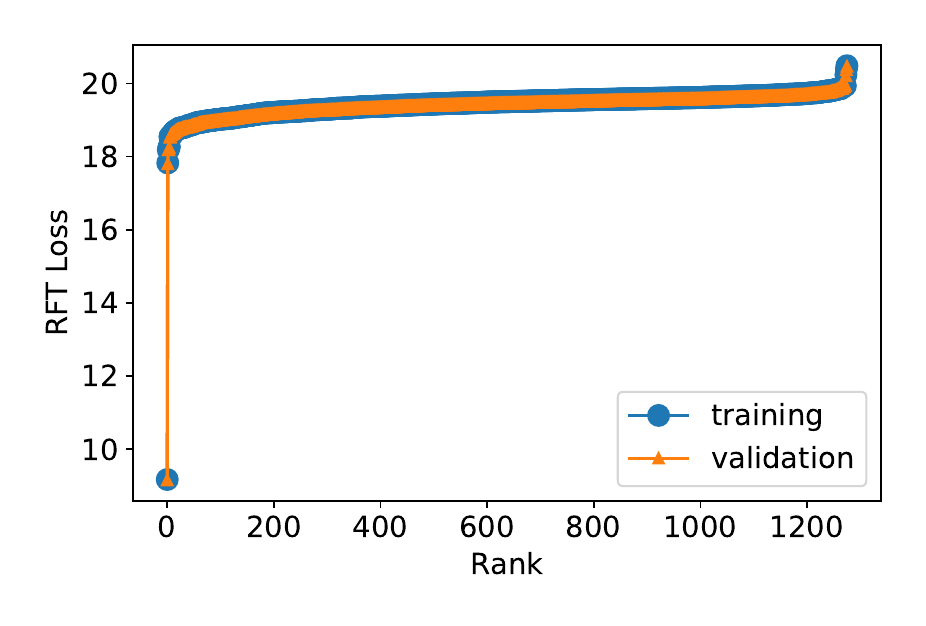}}
\caption{The sorted RFT loss for the raw representations obtained from unsupervised representation learning at resolution Level 1.}
\label{fig:feat_sorted_rft}
\end{figure}

\begin{figure}[t]
\centering{\includegraphics[width=0.7\linewidth]{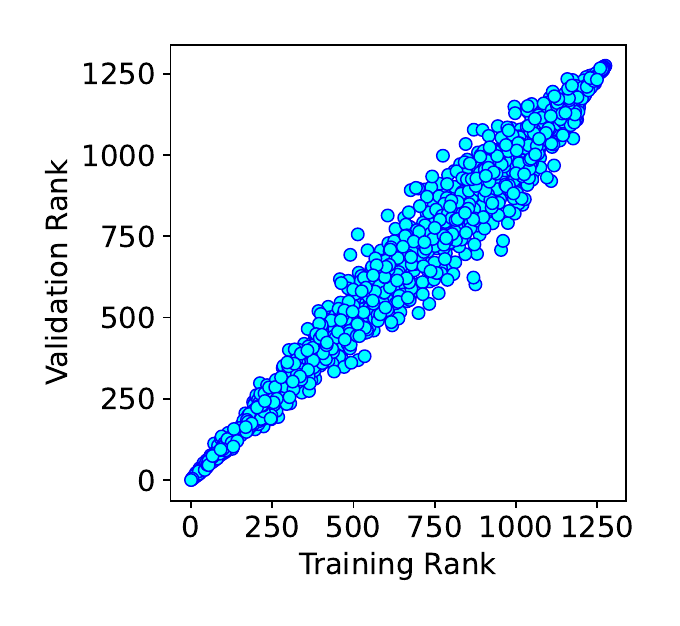}}
\caption{The training-validation joint RFT ranking for the raw representations obtained from unsupervised representation learning at resolution Level 1.}
\label{fig:feat_joint_ranking}
\end{figure}

\begin{figure}[t]
\centering{\includegraphics[width=0.85\linewidth]{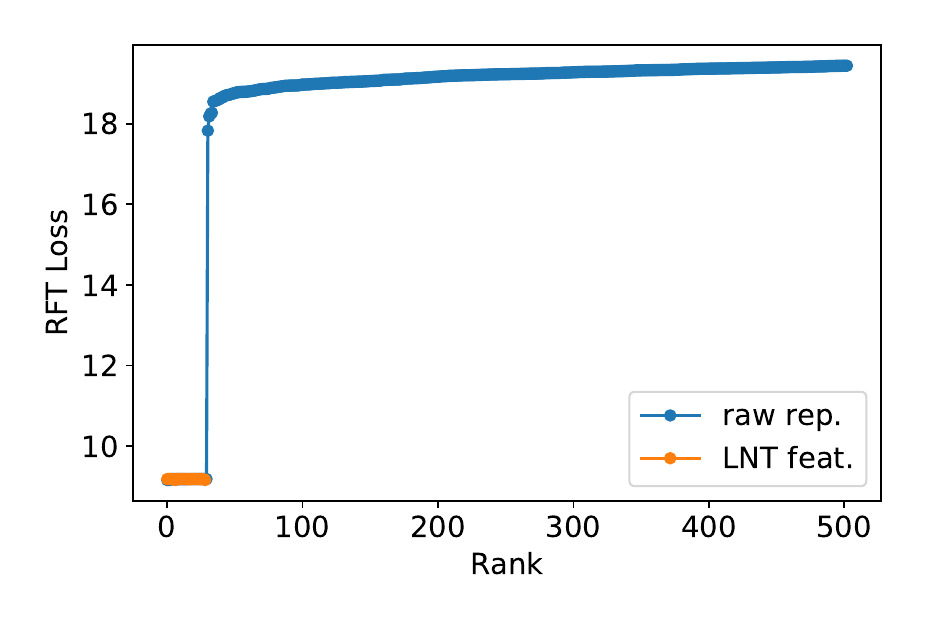}}
\caption{The sorted RFT loss of LNT features at resolution Level 1 as compared to the raw representations selected by RFT in Fig. \ref{fig:feat_sorted_rft}.}
\label{fig:feat_sorted_rft_lnt}
\end{figure}

\section{EXPERIMENTS} \label{sec:experiment}

In this section, we present a series of experiments to evaluate our model and compare its performance with DL models for LDCT image restoration. First, we describe the dataset and the corresponding GUSL setting. Next, we will show the restoration performance and computational efficiency of GUSL by performance benchmarking. Finally, we will illuminate our algorithm by discussing the intermediate and final experimental results.

\subsection{Experimental Setup} 

To evaluate the restoration performance and model efficiency of GUSL against deep learning models, we follow the conventional routine by adopting the \textit{2016 NIH-AAPM-Mayo Clinic LDCT Grand Challenge} \cite{mccollough2017low} dataset for model training and testing. This publicly released dataset comprises two types of images in pairs: full-dose NDCT images acquired at 120 kV and 200 quality-referenced effective mAs, and quarter-dose LDCT images obtained at 120 kV and 50 quality-referenced mAs. The entire dataset contains 2,378 NDCT-LDCT image pairs from 10 patients. Each image has a resolution of 512 × 512 pixels and a layer thickness of 3 mm, reconstructed using a B30 reconstruction kernel. In line with the experimental setting of \cite{wang2023ctformer}, data from patient L506 (211 testing images) is reserved for testing, while data from the remaining nine patients is employed for model training. We set the spatial resolution to $512 \times 512$, $256 \times 256$, $128 \times 128$, and $64 \times 64$ from Level 1 to Level 4 in the proposed GUSL pipeline.

\vspace{-3pt}
\subsection{Quality Performance Comparison}

We compare GUSL with five prominent LDCT restoration methods: RED-CNN \cite{chen2017low}, WGAN-VGG \cite{yang2018low}, MAP-NN \cite{shan2019competitive}, AD-NET \cite{tian2020attention}, and CTformer \cite{wang2023ctformer}. They respectively use multiple types of DL models. Two quality metrics, PSNR and SSIM, are employed to evaluate the restoration results. PSNR is useful for measuring the low-level pixel-wise restoration quality, while SSIM is suitable for measuring high-level image information such as structure, luminance, and contrast. The results are shown in Table \ref{tab:comparison}. GUSL achieves a PSNR value of 33.00 dB and an SSIM value of 0.9111. They are extremely close to the performance numbers obtained the top performer, CTformer. 

We show two source images and three restored images in Figs. \ref{fig:visualization}(a)-(e) for visual comparison in global and local views. GUSL and CTformer stand out among the three benchmarks. GUSL can productively restore the LDCT images against image noise while effectively preserving the global structures and the local details of images.

\subsection{Model Size and Complexity Comparison}

Table \ref{tab:comparison} compares the parameter count and the multiplier and addition counts (MACs) of the six models. GUSL has the smallest model size (0.57M parameters) and the lowest MACs (0.03M MACs/pixel). They are 39\% and 15\% of CTformer, respectively. This reduction demonstrates GUSL's efficiency, making it a compelling choice for a resource-constrained platform while still delivering state-of-the-art restoration performance.


\subsection{Feature Analysis} 

We analyze the properties of features obtained by unsupervised representation learning and supervised feature learning to illustrate their functions. 

Fig. \ref{fig:feat_sorted_rft} presents the quality of the features at resolution Level 1 of GUSL by showing the sorted RFT loss curve of the initial representation set from the best to the worst. The lower the RFT loss, the more powerful the feature. The apparent elbow point can effectively determine the number of representations to use in the later stages. 

Fig. \ref{fig:feat_joint_ranking} shows the joint RFT ranking for the training and validation data with the 80\% vs. 20\% split at resolution Level 1. Each dot indicates the ranking order of a feature dimension against the training data (along the horizontal axis) and the validation data (along the vertical axis).
We see most dots are distributed along the diagonal direction. The consistency in ranking distribution demonstrates the statistical stability of the raw representation set and the feasibility of effective representation selection by RFT. Representations located near the origin, which exhibit strong discriminative power across both datasets, are selected for the subsequent new feature generation step, thereby reducing the risk of overfitting.

Fig. \ref{fig:feat_sorted_rft_lnt} shows the RFT loss values of the newly generated LNT features at resolution Level 1. They are significantly smaller than the original representation data. This means that LNT features have superior discriminative capability over raw representations in the unsupervised representation learning module.

\section{Conclusion}\label{sec:conclusion}

In this work, we proposed a Green U-shaped Learning (GUSL) framework for low-dose CT (LDCT) image restoration. The new method is characterized by mathematical transparency, scalable model size, low power consumption, and high restoration performance. GUSL progressively predicts residuals from the coarsest level to the finest level, enabling an interpretable optimization process in contrast with the black-box, end-to-end optimized deep-learning solutions. Our design significantly reduces the size of the model and the complexity of inference of the learning systems. The effectiveness and efficiency of the proposed GUSL architecture were verified by experimental results. The compact model size and low MACs make GUSL suitable for implementations on mobile/edge devices while maintaining high-quality restoration performance. Furthermore, the explainable feature learning mechanism offers CT professionals insight into the imaging results.


\newpage
\bibliographystyle{IEEEbib}
\bibliography{refs}

@misc{pan2009computed,
  title={Computed tomography: From photon statistics to modern cone-beam CT},
  author={Pan, Tinsu},
  year={2009},
  publisher={Soc Nuclear Med}
}

@article{diwakar2018review,
  title={A review on CT image noise and its denoising},
  author={Diwakar, Manoj and Kumar, Manoj},
  journal={Biomedical Signal Processing and Control},
  volume={42},
  pages={73--88},
  year={2018},
  publisher={Elsevier}
}

@article{wang2006penalized,
  title={Penalized weighted least-squares approach to sinogram noise reduction and image reconstruction for low-dose X-ray computed tomography},
  author={Wang, Jing and Li, Tianfang and Lu, Hongbing and Liang, Zhengrong},
  journal={IEEE transactions on medical imaging},
  volume={25},
  number={10},
  pages={1272--1283},
  year={2006},
  publisher={IEEE}
}

@article{manduca2009projection,
  title={Projection space denoising with bilateral filtering and CT noise modeling for dose reduction in CT},
  author={Manduca, Armando and Yu, Lifeng and Trzasko, Joshua D and Khaylova, Natalia and Kofler, James M and McCollough, Cynthia M and Fletcher, Joel G},
  journal={Medical physics},
  volume={36},
  number={11},
  pages={4911--4919},
  year={2009},
  publisher={Wiley Online Library}
}

@article{li2014adaptive,
  title={Adaptive nonlocal means filtering based on local noise level for CT denoising},
  author={Li, Zhoubo and Yu, Lifeng and Trzasko, Joshua D and Lake, David S and Blezek, Daniel J and Fletcher, Joel G and McCollough, Cynthia H and Manduca, Armando},
  journal={Medical physics},
  volume={41},
  number={1},
  pages={011908},
  year={2014},
  publisher={Wiley Online Library}
}

@article{feruglio2010block,
  title={Block matching 3D random noise filtering for absorption optical projection tomography},
  author={Feruglio, P Fumene and Vinegoni, Claudio and Gros, J and Sbarbati, Andrea and Weissleder, RJPiM},
  journal={Physics in Medicine \& Biology},
  volume={55},
  number={18},
  pages={5401},
  year={2010},
  publisher={IOP Publishing}
}

@article{chen2017lowcnn,
  title={Low-dose CT via convolutional neural network},
  author={Chen, Hu and Zhang, Yi and Zhang, Weihua and Liao, Peixi and Li, Ke and Zhou, Jiliu and Wang, Ge},
  journal={Biomedical optics express},
  volume={8},
  number={2},
  pages={679--694},
  year={2017},
  publisher={Optica Publishing Group}
}

@article{chen2017low,
  title={Low-dose CT with a residual encoder-decoder convolutional neural network},
  author={Chen, Hu and Zhang, Yi and Kalra, Mannudeep K and Lin, Feng and Chen, Yang and Liao, Peixi and Zhou, Jiliu and Wang, Ge},
  journal={IEEE transactions on medical imaging},
  volume={36},
  number={12},
  pages={2524--2535},
  year={2017},
  publisher={IEEE}
}

@article{wolterink2017generative,
  title={Generative adversarial networks for noise reduction in low-dose CT},
  author={Wolterink, Jelmer M and Leiner, Tim and Viergever, Max A and I{\v{s}}gum, Ivana},
  journal={IEEE transactions on medical imaging},
  volume={36},
  number={12},
  pages={2536--2545},
  year={2017},
  publisher={IEEE}
}

@article{yang2018low,
  title={Low-dose CT image denoising using a generative adversarial network with Wasserstein distance and perceptual loss},
  author={Yang, Qingsong and Yan, Pingkun and Zhang, Yanbo and Yu, Hengyong and Shi, Yongyi and Mou, Xuanqin and Kalra, Mannudeep K and Zhang, Yi and Sun, Ling and Wang, Ge},
  journal={IEEE transactions on medical imaging},
  volume={37},
  number={6},
  pages={1348--1357},
  year={2018},
  publisher={IEEE}
}

@article{shan2019competitive,
  title={Competitive performance of a modularized deep neural network compared to commercial algorithms for low-dose CT image reconstruction},
  author={Shan, Hongming and Padole, Atul and Homayounieh, Fatemeh and Kruger, Uwe and Khera, Ruhani Doda and Nitiwarangkul, Chayanin and Kalra, Mannudeep K and Wang, Ge},
  journal={Nature Machine Intelligence},
  volume={1},
  number={6},
  pages={269--276},
  year={2019},
  publisher={Nature Publishing Group UK London}
}

@article{tian2020attention,
  title={Attention-guided CNN for image denoising},
  author={Tian, Chunwei and Xu, Yong and Li, Zuoyong and Zuo, Wangmeng and Fei, Lunke and Liu, Hong},
  journal={Neural Networks},
  volume={124},
  pages={117--129},
  year={2020},
  publisher={Elsevier}
}

@article{wang2023ctformer,
  title={CTformer: convolution-free Token2Token dilated vision transformer for low-dose CT denoising},
  author={Wang, Dayang and Fan, Fenglei and Wu, Zhan and Liu, Rui and Wang, Fei and Yu, Hengyong},
  journal={Physics in Medicine \& Biology},
  volume={68},
  number={6},
  pages={065012},
  year={2023},
  publisher={IOP Publishing}
}

@article{kuo2023green,
  title={Green learning: Introduction, examples and outlook},
  author={Kuo, C-C Jay and Madni, Azad M},
  journal={Journal of Visual Communication and Image Representation},
  volume={90},
  pages={103685},
  year={2023},
  publisher={Elsevier}
}

@article{yang2022supervised,
  title={On supervised feature selection from high dimensional feature spaces},
  author={Yang, Yijing and Wang, Wei and Fu, Hongyu and Kuo, C-C Jay and others},
  journal={APSIPA Transactions on Signal and Information Processing},
  volume={11},
  number={1},
  year={2022},
  publisher={Now Publishers, Inc.}
}

@inproceedings{wang2024statistics,
  title={A Statistics-based Feature Generation (SFG) Method: Theory and Applications},
  author={Wang, Xinyu and Wu, Yixing and Li, Haiyi and Mishra, Vinod K and Kuo, C-C Jay},
  booktitle={2024 IEEE International Conference on Big Data (BigData)},
  pages={5731--5738},
  year={2024},
  organization={IEEE}
}

@inproceedings{chen2016xgboost,
  title={Xgboost: A scalable tree boosting system},
  author={Chen, Tianqi and Guestrin, Carlos},
  booktitle={Proceedings of the 22nd acm sigkdd international conference on knowledge discovery and data mining},
  pages={785--794},
  year={2016}
}

@article{mccollough2017low,
  title={Low-dose CT for the detection and classification of metastatic liver lesions: results of the 2016 low dose CT grand challenge},
  author={McCollough, Cynthia H and Bartley, Adam C and Carter, Rickey E and Chen, Baiyu and Drees, Tammy A and Edwards, Phillip and Holmes III, David R and Huang, Alice E and Khan, Farhana and Leng, Shuai and others},
  journal={Medical physics},
  volume={44},
  number={10},
  pages={e339--e352},
  year={2017},
  publisher={Wiley Online Library}
}

@inproceedings{ronneberger2015u,
  title={U-net: Convolutional networks for biomedical image segmentation},
  author={Ronneberger, Olaf and Fischer, Philipp and Brox, Thomas},
  booktitle={Medical image computing and computer-assisted intervention--MICCAI 2015: 18th international conference, Munich, Germany, October 5-9, 2015, proceedings, part III 18},
  pages={234--241},
  year={2015},
  organization={Springer}
}

\end{document}